\documentclass[conference]{IEEEtran}
\IEEEoverridecommandlockouts
\usepackage{cite}
\usepackage{amsmath,amssymb,amsfonts}
\usepackage{algorithmic}
\usepackage{graphicx}
\usepackage{textcomp}
\usepackage{xcolor}
\usepackage{eso-pic}
\usepackage{multirow}
\def\BibTeX{{\rm B\kern-.05em{\sc i\kern-.025em b}\kern-.08em
    T\kern-.1667em\lower.7ex\hbox{E}\kern-.125emX}}
\begin{document}

\title{Frame-to-Panorama Localization and Context-Aware Sampling for Scene-Specific Ship Detection in a Smart Marina Testbed
\thanks{This work was co-financed by the European Union—NextGenerationEU,
through the Research and Innovation Foundation with grant number
STRATEGIC INFRASTRUCTURES/1222/0113 (MDigi-I), and the EU H2020
Research and Innovation Programmes under Grant Agreements No. 857586
(CMMI-MaRITeC-X).}
}

\author{\IEEEauthorblockN{1\textsuperscript{st} Ignat Romanov}
\IEEEauthorblockA{\textit{Department of Computer Science} \\
\textit{University of Nicosia}\\
Nicosia, Cyprus \\
romanov.i@live.unic.ac.cy}
\and
\IEEEauthorblockN{2\textsuperscript{nd} Andreas Hadjipieris}
\IEEEauthorblockA{\textit{Maritime Digitalization Centre} \\
\textit{Cyprus Marine and Maritime Institute}\\
Larnaca, Cyprus \\
andreas.hadjipieris@cmmi.blue}
\and
\IEEEauthorblockN{3\textsuperscript{rd} Neofytos Dimitriou}
\IEEEauthorblockA{\textit{Maritime Digitalization Centre} \\
\textit{Cyprus Marine and Maritime Institute}\\
Larnaca, Cyprus \\
neofytosd@gmail.com}

}

\maketitle
\AddToShipoutPictureFG*{%
  \AtPageLowerLeft{%
    \raisebox{4mm}{%
      \makebox[\paperwidth][c]{%
        \parbox{0.92\textwidth}{%
          \centering
          \fontsize{6}{7}\selectfont
          \textcopyright{} 2026 IEEE. Personal use of this material is permitted.
          Permission from IEEE must be obtained for all other uses, in any current
          or future media, including reprinting/republishing this material for
          advertising or promotional purposes, creating new collective works,
          for resale or redistribution to servers or lists, or reuse of any
          copyrighted component of this work in other works.\\
        }%
      }%
    }%
  }%
}
\begin{abstract}
Smart maritime infrastructures provide continuous access to heterogeneous sensing streams, enabling repeated experimentation, digital-twin development, and AI-based maritime services. However, sensing hardware alone is not sufficient for scene-specific model development: historical video streams must also be spatially indexed, contextualized, and reduced to informative subsets for annotation. This paper presents a frame-to-panorama localization and context-aware sampling pipeline for ship detection in historical PTZ maritime video lacking reliable pan, tilt, and zoom metadata. The main contribution is an end-to-end data-curation approach that recovers camera-view information from historical PTZ video and combines it with environmental context and visual diversity to construct compact, scene-specific training sets. Specifically, frames are localized on a reference panorama using SuperPoint and LightGlue, enriched with weather and solar-state metadata, and selected through diversity sampling to preserve variation across camera view and environmental conditions. A second context-aware stage targets under-represented distant-vessel cases near the horizon using tile-level visual embeddings and Gaussian Mixture Model clustering. Applied within the CMMI MDigi-I Smart Marina testbed, the proposed pipeline reduces 40,718 candidate frames to 220 images for annotation, corresponding to a 99.5\% reduction. A YOLO26-m detector fine-tuned on this subset achieves a mean AP50 of 94.78\% $\pm$ 0.51\% and a mean AP50--95 of 75.10\% $\pm$ 1.73\% under sequence-grouped five-fold cross-validation. These results demonstrate that highly redundant infrastructure video streams can be transformed into compact, spatially and contextually diverse training sets for scene-specific detector adaptation while substantially reducing annotation effort.
\end{abstract}

\begin{IEEEkeywords}
maritime surveillance, PTZ camera, ship detection, frame sampling, keyframe extraction, SuperPoint, LightGlue, object detection
\end{IEEEkeywords}

\section{Introduction}


\begin{figure*}[htbp]
\centering
\includegraphics[width=1\linewidth]{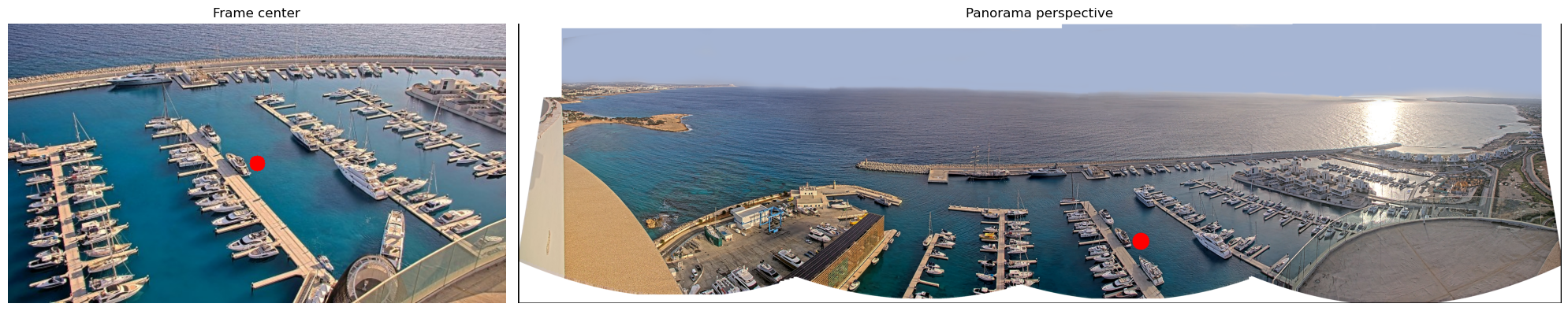}
\caption{Example of a video frame on the left. The center of this example is mapped to the panorama perspective through a homography transformation as shown on the right.}
\label{fig:centerpano}
\end{figure*}

\begin{figure*}[htbp]
\centering
\includegraphics[width=1\linewidth]{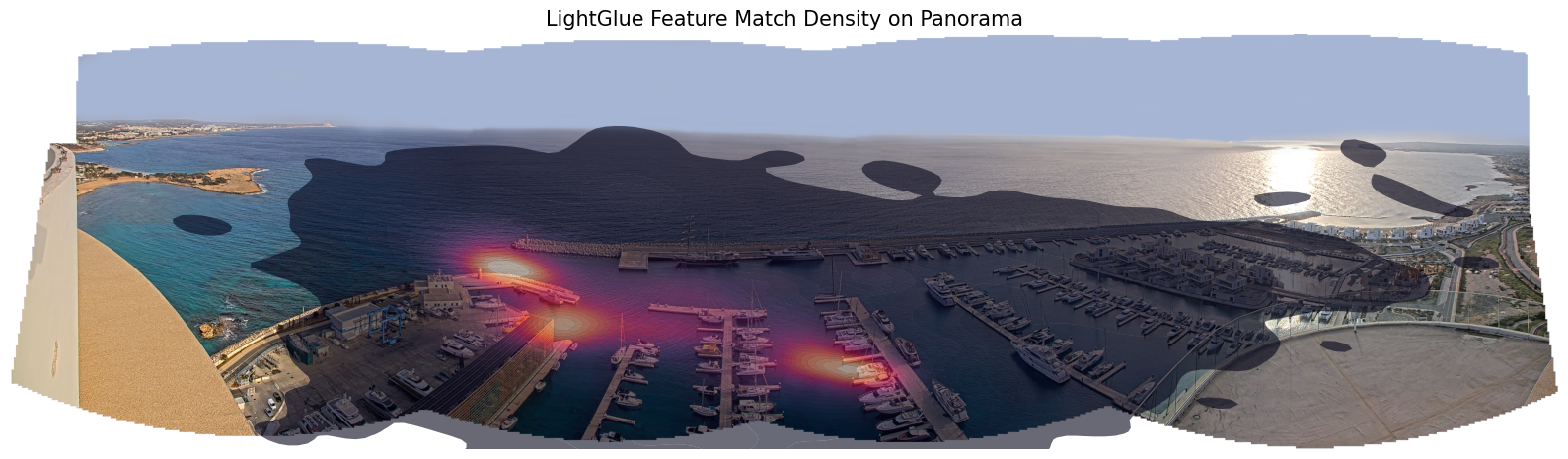}
\caption{Spatial distribution of mapped video frames on the reference panorama. The heatmap highlights frequently observed camera poses, including the dominant default park position.}
\label{fig:density}
\end{figure*}

Modern maritime research infrastructures are increasingly designed as operational testbeds where permanently stationed sensing assets support data collection, pilot execution, digital twinning, and the development of AI-enabled maritime services. Unlike short-term measurement campaigns, such infrastructures provide continuous access to heterogeneous data streams under real environmental and operational conditions. This lowers the barrier for developing and validating computer-vision methods, since models can be trained and tested on data collected from the same or similar physical environment in which they are expected to operate.

The data used in this work were collected within the CMMI MDigi-I Smart Marina infrastructure at Ayia Napa Marina, Cyprus. The Smart Marina integrates RGB and thermal Pan-Tilt-Zoom (PTZ) cameras, LiDAR, underwater acoustic sensors, weather and air-quality monitoring nodes, and edge-to-cloud computational resources. The camera considered in this work is installed approximately $100$ m above the water, providing a high-altitude view of the marina and surrounding maritime area, as illustrated in Fig.~\ref{fig:centerpano}. This setting provides a realistic testbed for scene-specific ship detection, with repeated camera poses, changing illumination, adverse weather, sea glare, variable zoom levels, and vessels appearing at very different scales. However, sensing hardware alone is not sufficient for adapting ML-based detectors to a specific deployment environment. The continuous PTZ video stream must first be converted into a compact and informative training set, which typically requires ground-truth annotation. Exhaustively annotating the available frames was impractical within the timeline of this work and would also be inefficient, since many sampled frames were highly redundant due to repeated camera poses and visually similar scene content.

Although video-frame redundancy and efficient sampling have been explored in prior work~\cite{yoon2023exploring,jadon2020unsupervised,sinulingga2023keyframe}, less attention has been paid to practical infrastructure-specific pipelines that make redundant sensor streams ready for annotation and model adaptation when camera-view metadata are missing. In such settings, data reduction cannot rely only on uniform temporal subsampling. Instead, frames must be temporally indexed, enriched with contextual information, and related to the physical space observed by the sensors. This is particularly important for PTZ cameras, where visual content depends on view direction and zoom level. When reliable pan, tilt, and zoom values are unavailable, it becomes difficult to curate training and evaluation data sets that cover the full operational field of view. Similar challenges arise in other infrastructure settings involving movable or mobile sensing platforms, where visual observations must be mapped to a common spatial reference before they can support robust AI development or digital-twin integration.

In our installation, up to $90\%$ of the recorded footage originates from the camera's default park position, producing strong spatial redundancy in the available data, as shown in Fig.~\ref{fig:density}. Random frame sampling would therefore over-represent this dominant view and under-represent less frequent camera poses or environmental conditions. To make the infrastructure data usable for efficient annotation and model training, the available frames must first be mapped to a common spatial reference and then sampled according to both camera-view diversity and environmental context.

To address this bottleneck, we propose a frame-to-panorama localization and sampling pipeline for historical maritime PTZ video without reliable pan, tilt, and zoom metadata. Individual frames are localized on a manually constructed reference panorama using SuperPoint~\cite{detone2018superpoint} and LightGlue~\cite{lindenberger2023lightglue}, yielding a panorama-coordinate representation of the camera view, including the projected frame center and an approximate view-scale measure derived from the projected frame coverage. These recovered view attributes are combined with weather observations and solar-state labels to support diversity sampling across camera view and environmental conditions. Because rare distant-vessel cases remained under-represented after this step, a second context-aware filtering stage isolates the horizon-facing maritime region, extracts tile-level visual embeddings, and uses Gaussian Mixture Model clustering to retain frames containing vessel-relevant open-water context. The resulting pipeline converts redundant smart-marina video streams into a compact subset for annotation and scene-specific detector training.

The contributions of this work are fourfold:
\begin{enumerate}
\item We demonstrate how a smart-marina infrastructure can be used as an AI model development testbed for scene-specific maritime perception.
\item We introduce and evaluate a frame-to-panorama localization pipeline that uses deep local features to map historical PTZ video frames to a reference panorama without relying on internal camera metadata.
\item We propose a sampling methodology that combines Kennard-Stone diversity sampling, weather and lighting metadata, and GMM-based context discovery for distant vessels near the horizon, reducing the data volume by $99.5\%$.
\item We manually annotate the selected subset of 220 images and use it to fine-tune an NMS-free YOLO26-m detector. Sequence-grouped five-fold outer cross-validation yields 94.78\% ± 0.51\% AP50 and 75.10\% ± 1.73\% AP50–95.
\end{enumerate}

\section{Methods}

The original data set was collected from a PTZ CCTV camera installed approximately $100$ m above Ayia Napa Marina, Cyprus, providing a high-altitude view of the marina and surrounding maritime area. Recordings were collected between 1 August 2025 and 10 January 2026 and were automatically saved when activity was detected in the scene or when the camera was manually controlled by an operator. One frame was extracted per minute. The resulting frames did not include reliable pan, tilt, zoom, weather, or lighting metadata. We note that in this work we used data frames recorded during the daylight (based on the solar telemetry metadata we describe in following subsections) which constituted the largest portion of the original data ($40,718$ out of $64,000$ frames).

\subsection{Camera-view recovery}


To assign each frame a spatial reference, we constructed a panorama using Adobe Photoshop~\cite{adobe_photoshop_cc_2019} by manually stitching sequential, non-magnified frames covering the camera's operational field of view. This panorama served as the common coordinate system for frame-to-panorama localization. For each video frame, SuperPoint~\cite{detone2018superpoint} was used to extract local keypoints and descriptors from both the frame and the panorama, while LightGlue~\cite{lindenberger2023lightglue} was used to establish feature correspondences. A homography was then estimated from the matched points using USAC-MAGSAC~\cite{barath2020magsac++}, and the frame centre was projected onto the panorama coordinate system (an example is visualized in Figure~\ref{fig:centerpano}). This produced the projected frame centre and the projected frame coverage, with the latter used as an approximate view-scale measure (i.e.\ depth of magnification). For each localized frame, we also stored the mean and median LightGlue confidence, the number of geometric inliers, and the projected coverage area. Because the panorama is used for feature-based localization rather than pixel-level alignment, minor stitching seams and local geometric distortions can be tolerated provided that sufficient consistent correspondences remain in the overlapping region. This is supported by the robustness of SuperPoint features to homographic transformations~\cite{detone2018superpoint}, the demonstrated performance of SuperPoint--LightGlue for homography estimation under viewpoint and illumination changes~\cite{lindenberger2023lightglue}, and the use of USAC-MAGSAC to reject geometrically inconsistent correspondences during robust model estimation~\cite{barath2020magsac++}. Sensitivity nevertheless increases when highly zoomed frames overlap primarily with a stitching seam or locally distorted region, where too few consistent matches may remain for reliable homography estimation.
 Minor coordinate variations primarily affect frames close to spatial-bin boundaries and are unlikely to substantially alter the subsequent multi-feature diversity selection.
\begin{table*}[htbp]
\centering
\caption{Frame-to-panorama localization performance comparison between the proposed SuperPoint-LightGlue pipeline and the traditional SIFT-FLANN baseline. The task is evaluated as spatial-bin classification over the reference panorama.}

\label{tab:global_performance}
\begin{tabular}{|l|c|c|c|c|c|c|c|}
\hline
\multirow{2}{*}{\textbf{Algorithm}} 
& \multirow{2}{*}{\textbf{Accuracy}} 
& \multicolumn{3}{c|}{\textbf{Macro Average}} 
& \multicolumn{3}{c|}{\textbf{Weighted Average}} \\ 
\cline{3-8} 
& 
& \textbf{Precision} 
& \textbf{Recall} 
& \textbf{$F_1$-Score} 
& \textbf{Precision} 
& \textbf{Recall} 
& \textbf{$F_1$-Score} \\ 
\hline
\hline
\textbf{SuperPoint+LightGlue} 
& \textbf{87.36\%} 
& \textbf{62.69\%} 
& \textbf{68.82\%} 
& \textbf{62.83\%} 
& \textbf{93.67\%} 
& \textbf{87.36\%} 
& \textbf{89.93\%} \\
\hline
\textbf{SIFT+FLANN} 
& 12.34\% 
& 7.29\% 
& 7.03\% 
& 6.39\% 
& 12.77\% 
& 12.34\% 
& 11.22\% \\ 
\hline
\end{tabular}

\end{table*}

\subsection{Metadata enrichment and diversity sampling}

To enrich each image with environmental context, we integrate historical weather data and solar telemetry. For each image frame, the corresponding metadata includes the nearest hourly weather observation, provided that the weather timestamp is within 1 hour of the image timestamp, together with a solar-state label. The solar state is categorized as day, night, or transition based on the image timestamp relative to sunrise and sunset. A 30-minute buffer is applied around both sunrise and sunset: timestamps clearly between sunrise and sunset are labelled as day, timestamps clearly outside this interval are labelled as night, and timestamps falling within the buffer periods are labelled as transition. The transition category therefore captures the periods during which illumination changes from night to daylight or from daylight to night.

The enriched frame pool was then sampled to reduce redundancy while preserving diversity across camera pose and environmental conditions. First, frames were stratified into equal-width bins along the recovered horizontal panorama coordinate. From each bin, up to $n$ images were selected. If a bin contained fewer than $n$ images, all images were retained. Otherwise, the Kennard-Stone algorithm~\cite{kennard1969computer,kennardstonegit} was applied within the bin to Z-score normalized features to select a representative subset in a feature space defined by weather, vertical panorama coordinate, and projected frame coverage. This stage reduced the localized candidate set from $40,718$ to $2,402$ images, corresponding to a $94.1\%$ reduction.

\subsection{Context-aware sampling}
\begin{figure}[htbp]
\centering
\includegraphics[width=1\linewidth]{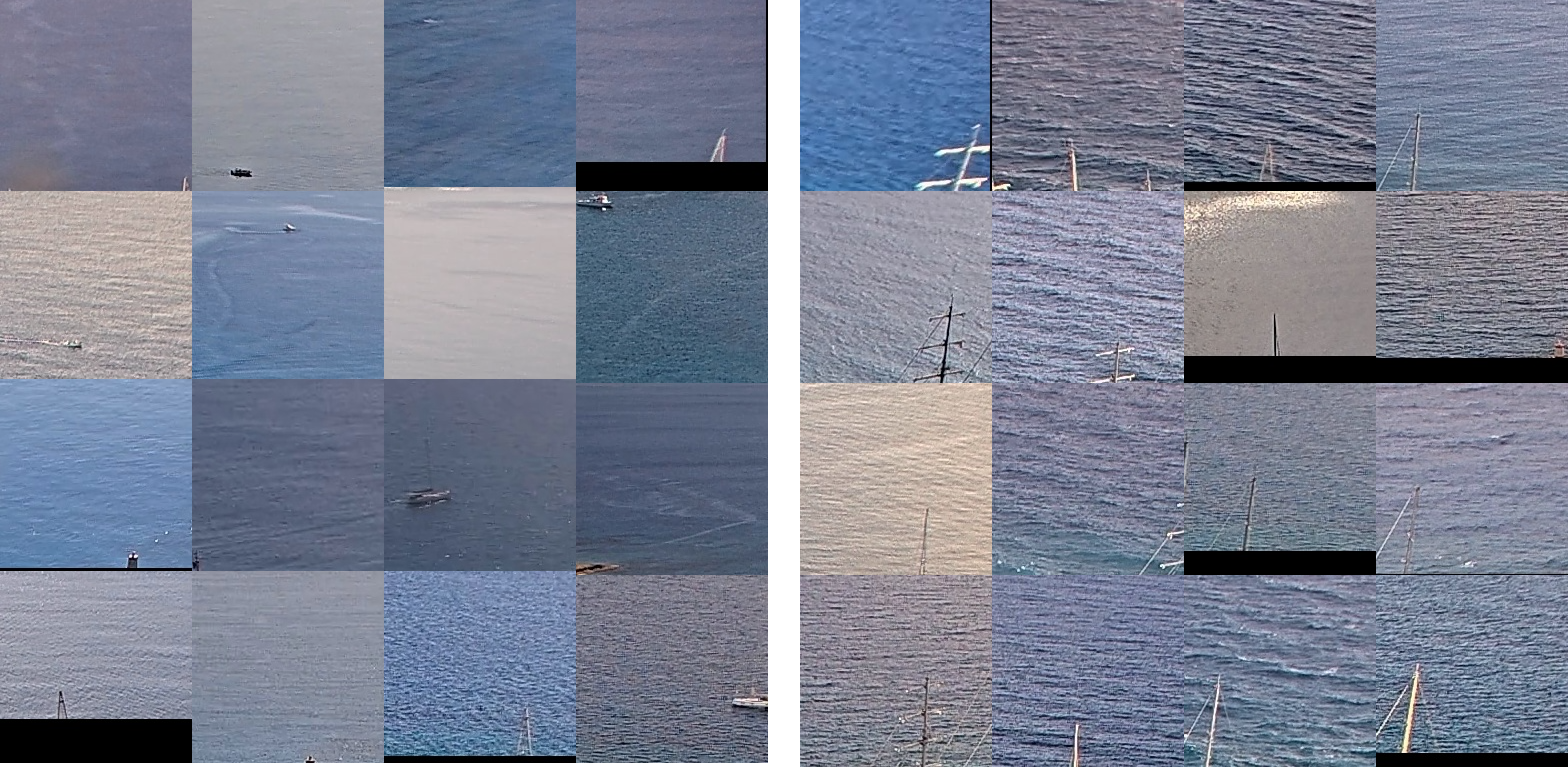}
\caption{Examples of GMM clusters obtained from tile-level visual embeddings. The left cluster contains vessel-relevant tiles, while the right cluster contains visually similar but non-relevant structures such as waves, masts, and breakwater regions.}
\label{fig:nonrelevantgmm}
\end{figure}

Following diversity sampling, a context-aware filtering step was applied to prioritize frames containing small maritime vessels in the open-water horizon outside the marina. Because frames were captured from different viewing directions and zoom levels, a lightweight YOLO11-n detector was trained to identify the horizon-facing maritime region, and each frame was cropped to this region of interest.

To preserve detail for distant-vessel discovery, each cropped region was divided into overlapping $192 \times 192$ pixel tiles with $10\%$ overlap in both dimensions, retaining edge tiles with a minimum size of $64$ pixels. Tile-level predictions and visual embeddings were extracted using an ATSS-SwinL-DyHead model, which was found to produce features more sensitive to small distant boats than YOLO11 during preliminary experiments. Since the model showed high recall but low precision, only tiles predicted as boat were retained for clustering. Their embeddings were reduced with PCA while preserving $90\%$ of the variance, and a full-covariance Gaussian Mixture Model was fitted to the reduced feature space, with the number of components selected using the Bayesian Information Criterion. Frames were retained if at least one tile belonged to a vessel-relevant cluster, allowing visually similar non-relevant structures such as waves, buoys, masts, and breakwater regions to be filtered out.  Figure \ref{fig:nonrelevantgmm} illustrates examples of GMM clusters corresponding to vessel-relevant tiles and visually similar non-relevant structures such as waves, masts, and breakwater regions. This stage further reduced the data set from $2,402$ to $220$ images.

\subsection{Annotation and model training}
\begin{figure}[htbp]
    \centering
    \includegraphics[width=1\linewidth]{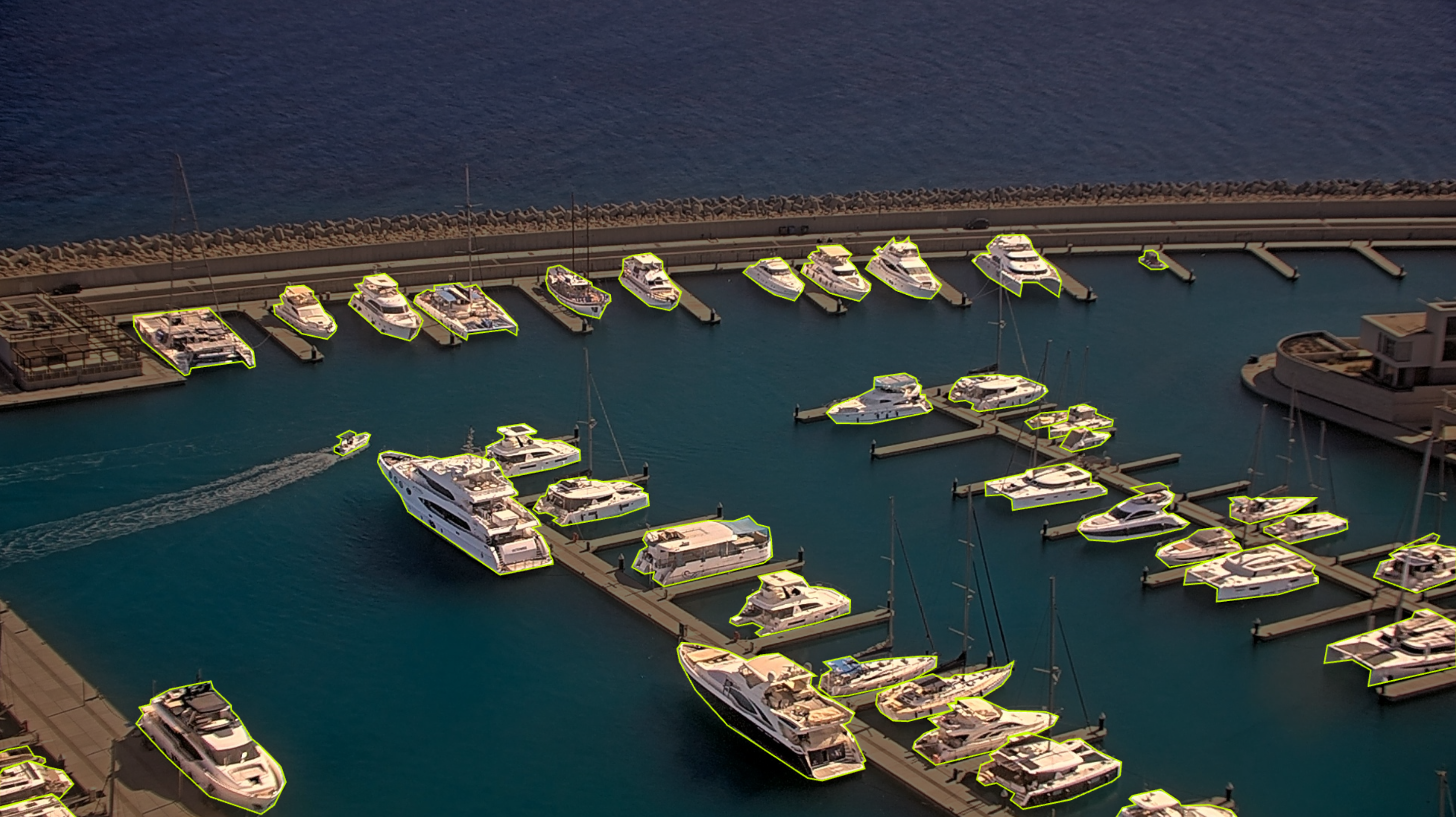}
    \caption{Example of a manually curated segmentation annotation used to create object-detection labels. The annotation protocol focuses on visible vessel structures while excluding thin or ambiguous elements such as masts and railings.}
    \label{fig:masksample1}
\end{figure}

The final dataset comprised 220 images containing 4,017 vessel bounding boxes; 208 images contained at least one labeled vessel and 12 contained no labeled vessels. Performance was evaluated using YOLO26-m~\cite{jocher2026ultralytics}, the medium-sized variant of the Ultralytics YOLO26 object-detection architecture, comprising 20.4 million parameters. The standard YOLO26-m architecture was used without architectural modifications. Model performance was evaluated using five-fold outer cross-validation. To prevent leakage between temporally related frames, the images were grouped into 182 acquisition sequences derived from their recording identifiers, and all frames from the same sequence were assigned to the same fold. Each outer fold contained 44 images. In each run, 158 images were used for training, 18 for internal validation and checkpoint selection, and 44 exclusively for outer-fold evaluation. Consequently, every image was evaluated exactly once by a model that had not used it for training or model selection. 
Each fold was initialized from the same pretrained YOLO26-m checkpoint and trained for up to 500 epochs at an input resolution of 1280 × 1280 pixels, using a batch size of 16, cosine learning-rate scheduling, automatic mixed precision, and early stopping with a patience of 50 epochs. The data-partition and training seeds were fixed at 42, and all folds used identical hyperparameters. The checkpoint with the best internal-validation performance was evaluated on the corresponding outer fold.

\section{Results \& Discussion}

\subsection{Frame-to-panorama localization}

To quantitatively and qualitatively evaluate the mapping between image frames and the reference panorama, we compared our method against a traditional baseline. The proposed pipeline uses SuperPoint features paired with LightGlue matching, while the traditional baseline relies on SIFT features matched via the Fast Library for Approximate Nearest Neighbors (FLANN).

To establish a quantitative benchmark, evaluation was conducted on video frames captured during daytime conditions. The evaluation set was structured by partitioning the horizontal panorama coordinate into $20$ equal-width spatial bins. To ensure a balanced distribution, up to $100$ images were sampled from each bin. Ground-truth labels were assigned manually by cross-referencing each sampled frame with the corresponding bin range on the reference panorama. This formulation allowed frame-to-panorama localization to be evaluated as a discrete spatial-bin classification problem.

As summarized in Table~\ref{tab:global_performance}, the SuperPoint+LightGlue pipeline significantly outperforms the traditional SIFT+FLANN baseline across all metrics, achieving a global accuracy of $87.36\%$ compared to the baseline's $12.34\%$.

\begin{figure*}[htp]
    \centering
    \includegraphics[width=1\linewidth]{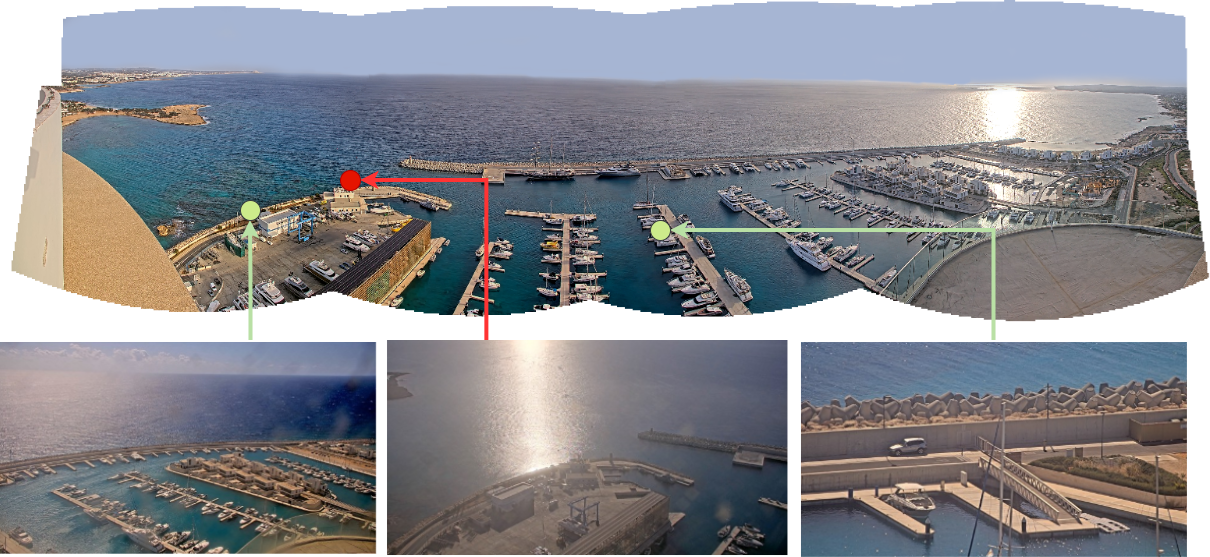}
    \caption{Representative frame-to-panorama localization success and failure cases. Green lines show SIFT-FLANN results and red lines show SuperPoint-LightGlue results. The dots indicate recovered panorama locations, while markers outside the panorama indicate failed localization cases where not enough reliable keypoints were detected. The bottom images show examples where one of the methods failed under challenging visual conditions.}
    \label{fig:deep_0_sift_wrong}
\end{figure*}

A qualitative analysis of the failure modes reveals distinct behaviors between the two approaches. The deep learning pipeline predominantly fails on highly zoomed-in frames. In these scenarios, the limited field of view reduces the number of identifiable keypoints, preventing the robust estimation of a homography matrix between the frame and the reference panorama. Conversely, under these identical conditions, the traditional SIFT+FLANN approach frequently produces a match; however, these matches are typically inaccurate, as shown in the top row of Figure~\ref{fig:deep_0_sift_wrong}. 


\begin{table*}[!t]
\centering
\caption{Sequence-grouped five-fold outer cross-validation performance of the
fine-tuned YOLO26-m detector. Each outer evaluation fold contains 44 images.
The final row reports the unweighted mean $\pm$ sample standard deviation
across folds.}
\label{tab:detection_performance}
\begin{tabular}{|l|c|c|c|c|c|}
\hline
\textbf{Fold}
& \textbf{Precision}
& \textbf{Recall}
& \textbf{F1-score}
& $\mathbf{AP_{50}}$
& $\mathbf{AP_{50\textnormal{--}95}}$ \\
\hline
\hline
Fold 1
& 0.9438
& 0.9015
& 0.9222
& 0.9485
& 0.7485 \\
\hline
Fold 2
& 0.9599
& 0.8928
& 0.9252
& 0.9500
& 0.7623 \\
\hline
Fold 3
& 0.9339
& 0.9092
& 0.9214
& 0.9527
& 0.7647 \\
\hline
Fold 4
& 0.9532
& 0.8738
& 0.9117
& 0.9392
& 0.7221 \\
\hline
Fold 5
& 0.9566
& 0.9067
& 0.9310
& 0.9486
& 0.7574 \\
\hline
\hline
\textbf{Mean $\boldsymbol{\pm}$ SD}
& $\mathbf{0.9495 \pm 0.0106}$
& $\mathbf{0.8968 \pm 0.0143}$
& $\mathbf{0.9223 \pm 0.0070}$
& $\mathbf{0.9478 \pm 0.0051}$
& $\mathbf{0.7510 \pm 0.0173}$ \\
\hline
\end{tabular}
\end{table*}
In contrast, instances where the deep learning model fails on zoomed-out frames are rare.


Finally, the middle example of Figure~\ref{fig:deep_0_sift_wrong} highlights a common scenario where the deep learning pipeline successfully registers the frame but the traditional pipeline fails. These cases are primarily attributed to extreme illumination discrepancies between the query frame and the reference panorama, showcasing the robust feature representation capabilities of the learning-based approach over handcrafted descriptors under variable lighting conditions.

\begin{figure}[htbp]
    \centering
    \includegraphics[width=1\linewidth]{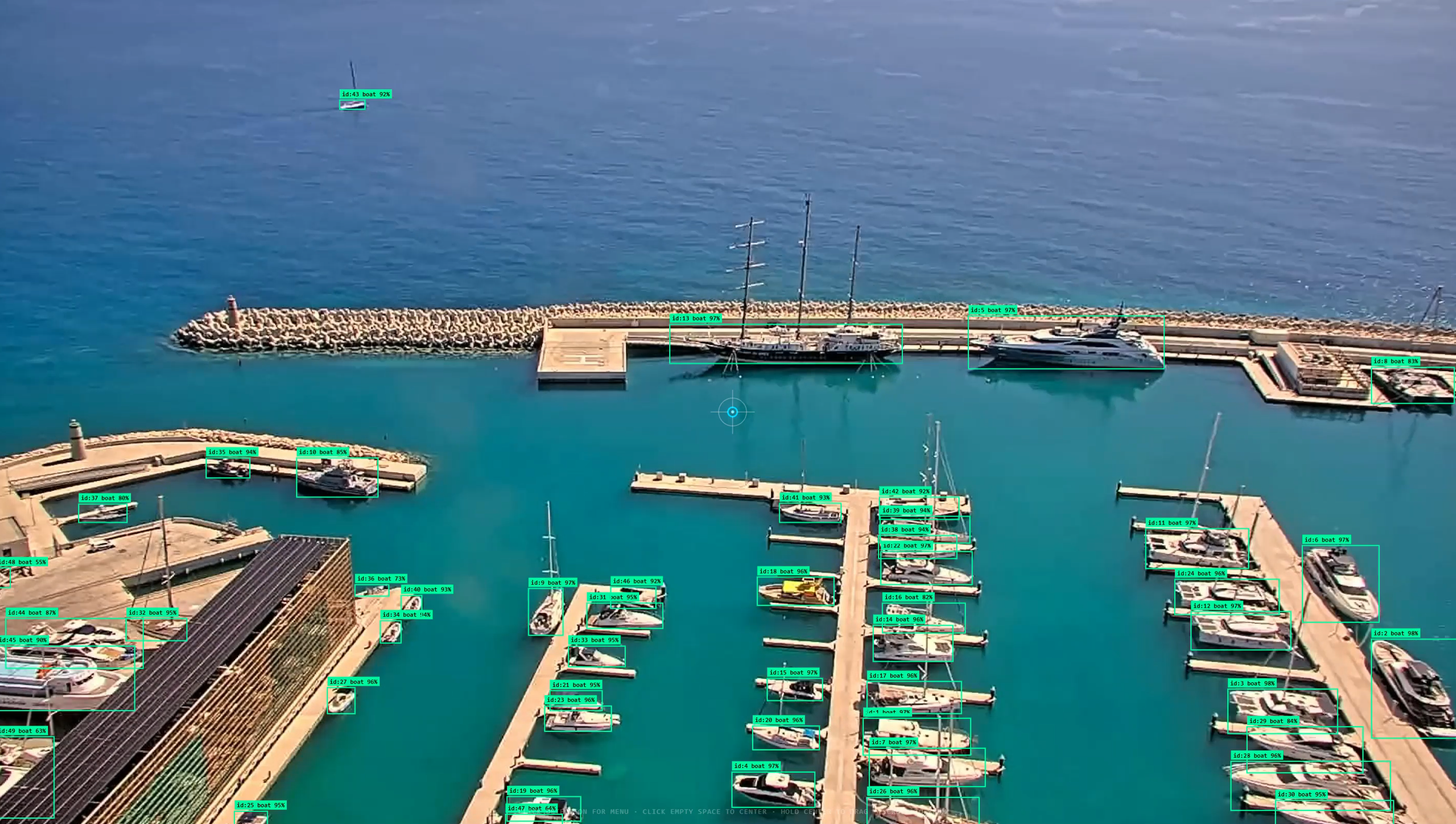}
    \caption{Example prediction from the live PTZ camera feed using the trained YOLO26-m detector. All boats within the marina are detected as well as a small boat on the horizon outside the marina.}
    \label{fig:modelpred}
\end{figure}

\subsection{Detection performance}

Table~\ref{tab:detection_performance} summarizes the sequence-grouped five-fold outer cross-validation results.
Fig.~\ref{fig:modelpred} shows an example prediction from the live PTZ camera feed, including detections within the marina and a small vessel near the horizon.

Across the five outer folds, AP50 was 94.78\% ± 0.51\% and AP50–95 was 75.10\% ± 1.73\% (mean ± sample SD). The fold-level AP50 values were 94.85\%, 95.00\%, 95.27\%, 93.92\%, and 94.86\%, corresponding to a range of 93.92\%–95.27\%. The two-sided 95\% fold-level Student-t interval for mean AP50 was 94.15\%–95.41\%. Mean precision, recall, and F1-score were 94.95\% ± 1.06\%, 89.68\% ± 1.43\%, and 92.23\% ± 0.70\%, respectively. The small AP50 standard deviation and narrow range indicate that performance was stable across partitions and was not attributable to a fortunate 90\%/10\% split.

The proposed pipeline enabled the selection of a compact annotation set from a large and redundant infrastructure video stream. The resulting $220$ annotated images captured different parts of the marina under varying camera views and environmental conditions, supporting scene-specific detector adaptation with substantially reduced annotation effort. This demonstrates how the Smart Marina infrastructure can be converted from a continuous sensing environment into an actionable AI model development testbed.

The current trained model sometimes fails to detect larger vessels that were not present in the training set, vessels under extreme sun glare, and ships in highly zoomed-in frames where the vessel occupies $\geq 50\%$ of the image and is truncated by the frame boundary. These limitations can be addressed by including more examples with extreme zoom levels, stronger glare, truncated vessels, and a wider range of vessel types and sizes. Future work will use these observed failure cases to guide subsequent data collection and annotation rounds within the Smart Marina infrastructure. Generalizability across marina-like environments will also be explored.




Future work will use the observed failure cases to guide subsequent data collection and annotation rounds within the Smart Marina infrastructure. In addition, the current manual Photoshop-based image-stitching step will be replaced by automated image-stitching frameworks, reducing manual intervention and improving the scalability and reproducibility of the data preparation pipeline. Moreover, the generalizability of the proposed approach across other marina-like environments will be explored.

\bibliographystyle{abbrv}
\bibliography{bib}

\end{document}